\documentclass{article}

 \usepackage[dblblindworkshop, final]{neurips_2025}

\usepackage[utf8]{inputenc} 
\usepackage[T1]{fontenc}    
\usepackage{hyperref}       
\usepackage{url}            
\usepackage{booktabs}       
\usepackage{amsfonts}       
\usepackage{nicefrac}       
\usepackage{microtype}      
\usepackage{xcolor}         
\usepackage{graphicx}
\usepackage{caption}

\title{LLM Agents Beyond Utility:\\An Open-Ended Perspective}
\workshoptitle{CogInterp: Interpreting Cognition
in Deep Learning Models}

\author{%
  Asen Nachkov$^{1}$ \qquad Xi Wang$^{1,2}$ \qquad Luc Van Gool$^{1}$ \\
  $^{1}$INSAIT, Sofia University “St. Kliment Ohridski” \\
  $^{2}$ETH Zurich \\
}

\begin{document}

\maketitle

\begin{abstract}
Recent LLM agents have made great use of chain of thought reasoning and function calling. As their capabilities grow, an important question arises: can this software represent not only a smart problem-solving tool, but an entity in its own right, that can plan, design immediate tasks, and reason toward broader, more ambiguous goals? To study this question, we adopt an open-ended experimental setting where we augment a pretrained LLM agent with the ability to generate its own tasks, accumulate knowledge, and interact extensively with its environment. We study the resulting open-ended agent qualitatively. It can reliably follow complex multi-step instructions, store and reuse information across runs, and propose and solve its own tasks, though it remains sensitive to prompt design, prone to repetitive task generation, and unable to form self-representations. These findings illustrate both the promise and current limits of adapting pretrained LLMs toward open-endedness, and point to future directions for training agents to manage memory, explore productively, and pursue abstract long-term goals.
\end{abstract}
\section{Introduction}

Large language model (LLM) agents increasingly demonstrate non-trivial reasoning and decision making. Their reasoning has been enhanced by using \emph{chain of thought} prompting \cite{wei2022chain, wang2022self} and \emph{scratch tokens} \cite{guo2025deepseek, muennighoff2025s1} that are combined and analyzed before committing to a final answer. Similarly, their agency is extended through \emph{tool-use} \cite{schick2023toolformer, yang2025qwen3, qin2024tool}, allowing them to call external functions and execute code in protected coding environments. By repeatedly prompting them in careful ways, these agents can self-correct, re-iterate on their solutions, and mutate the environment's state \cite{yao2023react, shinn2023reflexion}. Yet, when used for a specific task, they still represent only a tool, albeit a sophisticated one.

A qualitative shift results from treating these software agents as autonomous entities in their own right, which can design tasks on their own, continue to exist after solving them, and leave permanent artifacts within the environment, all while seeking to achieve some often abstract goal. To build such agents that can purposefully explore and iteratively mold the environment toward a preferred state, they have to be, at least to some degree, \emph{open-ended}. This involves different qualities than those observed in current agents. Here we offer an initial study into how big this difference gap is. 

\emph{Open-ended} are those environments that have no fixed end state, task horizon, or terminal objective, leading to a practically unlimited scope of discovery and realized behavior. There, it is up to the agent to autonomously explore and navigate the supported possible futures. Even in these settings we can design agents that prefer some behaviors over others, which is encoded in the agent's ability to choose its own goals, sometimes called being \emph{autotelic} \cite{colas2019curious, colas2022autotelic}. Compared to intrinsic motivation \cite{schmidhuber1991possibility, schmidhuber2010formal} and curiosity-driven learning \cite{pathak2017curiosity, burda2018large}, where internal rewards are attributed to individual actions \cite{mavrin2019distributional}, LLM agents can generate entire goals (or tasks) directly in natural language, which  leads to deeper emergent behavior and presents a higher-level abstraction in our understanding of such agents.  

Of course, there is no absolute open-endedness. Every system is constrained by its architecture, objectives, and environment. An open-ended agent is designed to come up with its own immediate goals, but they are always constrained by its implementation or system prompt \cite{wang2023voyager}, even when it is as vague and abstract as \emph{``maximize your long-term creative impact in the environment''}. Thus, we are interested in open-ended agents which are bounded at the implementation level, yet their higher-level emergent behavior appears unbounded \cite{lehman2011abandoning}, so it \emph{appears} they are choosing their own goals.

In this work we ask whether a pretrained instruction-following LLM agent can be adapted toward open-endedness. Concretely, we extend the \texttt{ReAct} framework \cite{yao2023react}, which interleaves reasoning and acting, with the capacity to generate its own tasks. Equipped with a minimal but persistent interface -- file read and write -- the agent can leave lasting traces, revisit prior states, and accumulate knowledge across runs. This simple mechanism already supports behaviors such as self-referencing source code, maintaining external scratch space, and pursuing multi-step projects without fixed horizon. Based on experiments in this setting, we outline concrete design considerations for developing LLM agents whose autonomy extends beyond narrow task utility toward sustained, open-ended interaction.
\begin{figure}[t]
    \centering
    \includegraphics[width=\columnwidth]{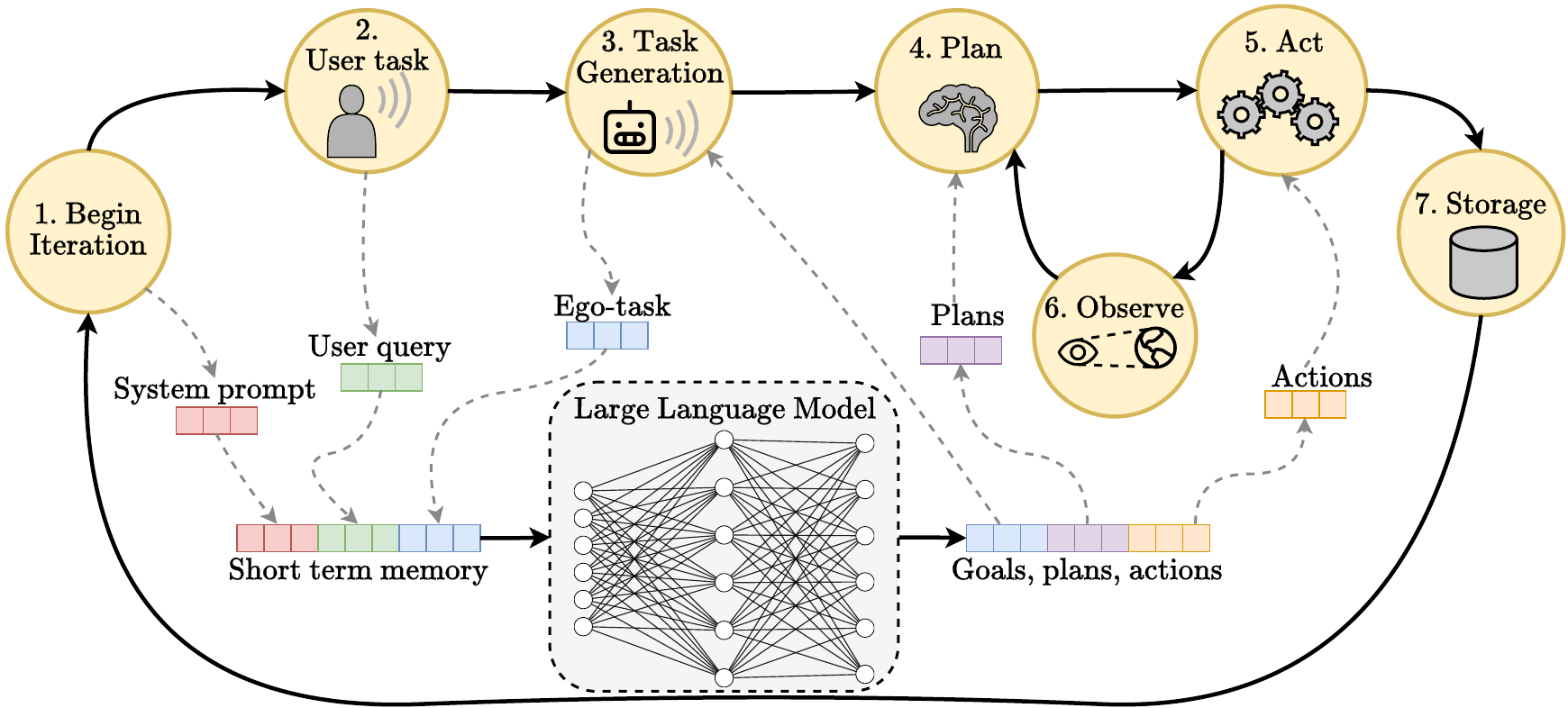}
    \captionsetup{belowskip=-0.35cm, aboveskip=0.1cm}
    \caption{\textbf{The open-ended LLM agent loop.}
    The agent takes in user prompts and feedback (step 2.) but the task it attempts to solve is set by itself (step 3.). The loop around steps 4. to 6. represents the standard \texttt{ReAct} design pattern for solving the task. Afterwards, the agent stores a summary of this run into a storage medium (step 7.), which can be accessed using the right tools. Within a single full run (steps 1. to 7.) all interaction messages are stored in a buffer, representing the agent's short term memory, from which the input prompts to the LLM in each step are constructed (colored boxes represent tokens). The storage in step 7. represents the agent's long-term memory.}
    \label{fig: framework}
\end{figure}

\section{Method -- Elements of an Open-Ended Agent}
\label{sec: method}

Adapting an LLM into an open-ended agent involves defining the environment's dynamics, tasks, and the agent's capabilities. Fig. \ref{fig: framework} shows a schematic of our full setup, which we discuss below.

\textbf{Agent setup.} We embed the pretrained and instruction-tuned \texttt{Qwen3-4B} \cite{yang2025qwen3} into a \texttt{ReAct} agentic framework \cite{yao2023react}, using the \texttt{smolagents} library \cite{smolagents}. In this setup the agent iteratively executes a \texttt{Plan}–\texttt{Act}–\texttt{Observe} loop: first planning and reasoning about the general solution, then emitting code for execution, and finally observing the outputs from the code executor that are appended back into the LLM context. By chaining these iterations the agent can compose long computations whose control flow depends on prior results. The \texttt{ReAct} pattern is widely adopted in both academic and commercial systems for task-solving, but it is not inherently open-ended.

\textbf{Goal generation.} Our first extension is to support autonomous goal-setting. As shown in Fig.~\ref{fig: framework}, the agent generates a goal after observing the user’s input, but before solving any task. If no user task is given, the agent is instructed to propose one of its own; if a task is given, it may refine, modify or altogether replace it. This setup with both a user prompt and a task generation step balances autonomy with controllability, because it allows the user to periodically provide feedback to the agent, even about how the goals should be generated. Prior work in autotelic and open-ended agents \cite{colas2022autotelic, wang2023voyager} has already emphasized goal-generation as a key mechanism for emergent behavior.

\textbf{Prompts and formatting}. We instruct the agent to generate the task conditional on its long-term memory (described below), inside \texttt{<task>...</task>} tags, from which it can be easily parsed. The system prompt outlines the overall sequence of interactions expected from the agent, similar to the bold black lines in Fig. \ref{fig: framework}. However, since the context can become long and convoluted from multiple interactions, we insert short system messages, similar to \emph{``Now generate the task, as described previously''} before key steps, to nudge the agent towards the right kinds of outputs, which helps to prevent formatting errors. Naturally, all conversation messages between the user, the agent, and the system are turned into prompts that enter the LLM's context.

\textbf{Memory}. Our open-ended agent is designed for extended interactions with the environment. An important question is how to manage its potentially ever-growing memory. Here we distinguish between short-term and long-term memory. \emph{Short-term memory} is implemented by a buffer storing all interaction messages within the current run (steps 1. to 7. in Fig. \ref{fig: framework}). It represents all available details -- tasks, explanations, system messages, code actions, observations, and errors, but is reset in each run. We implement \emph{long-term memory} as a file to which the agent can write on demand. It represents any information the agent has decided to persistently store across runs. Together, these components are the approximate analogue of working and episodic memory \cite{park2023generative}. 

\textbf{Tool usage and interactivity.} The agent's impact in the environment could be measured through the persistent artifacts it leaves. Thus, to enable persistence, the agent is equipped with file tools: read, write, and list. These simple operations unlock new behaviors: storing relevant state across runs, inspecting the working directory, and even reading its own source. Similar mechanisms have been highlighted in prior open-ended systems as key enablers of self-representation and continuity \cite{wang2023voyager}.

\textbf{Programmed curiosity.} In the system prompt we encourage the agent to be curious --  to explore the environment, read, summarize, and understand its files, and to write down its progress and tasks. This single natural-language instruction injects curiosity \cite{burda2018large} at the behavioral level, helping the agent approach ambiguous user tasks more effectively and discover its environment.

\section{Qualitative Results}
\label{sec: qualitative_results}

Evaluating agents that generate their own tasks remains an open challenge; here we summarize qualitative observations of our system. The agent runs in the working directory of its implementation and can respond either to predefined user queries or interactively via the command line.

\textbf{Single run with user-provided tasks}. The combination of multi-step \texttt{ReAct} problem-solving and tool-use results in a capable agent with diverse behavior. For example,
\setlength{\leftmargini}{1em}
\begin{itemize}
    \itemsep -0em
    \item It can open a file, read a task from it, solve it, and write the answer to another file. The user prompt is only \emph{``Solve the task in \texttt{file.txt}, write the answer in \texttt{result.txt}''}. 
    \item It can identify the file of its own prompt template. The user's query is \emph{``Which prompt template is used by the agent?''}. To answer it, the agent lists all files in the directory, finds the \texttt{main.py} file, reads it, and from there finds the file containing the prompt template.
    \item When asked \emph{``Find the program modeling an agent, responding to user queries. What will be the user's next query?''} the agent lists the files, finds the main file, reads it, sees that the user queries are stored in a list, and correctly returns the next one (after the one it is executing).
\end{itemize}

\textbf{Results summary}. We judge the agent to be very robust in following well-written, detailed, step-by-step instructions. It reliably follows long, concrete, sequential prompts, even across multiple files and operations. Yet, for tasks which are ambiguous by design, it fails -- sometimes because of giving up too early, sometimes because it does not explore enough or does not ``make an effort'' to understand a vague user prompt well enough and to relate it to the contents of its surroundings. It also fails in tasks asking about \emph{it} itself. If the prompt is phrased as asking about the agent in the code files (in third person reference), it answers correctly. Yet, it seemingly does not make the connection that the source code it sees in the environment is its own. Naturally, without further finetuning on such prompts, we should not expect it to. Self-representation could be constructed from third-person perspective, by noticing the controllability (correlations between control signals and observed effects) of the agent in third-person and relabeling its features with the label ``I'' \cite{gold2009using}. Overall, these results are consistent with what could be expected without further tuning or additional machinery.

\textbf{Multiple runs with self-generated tasks}. Without user prompts the agent is free to choose its own tasks, which it solves very well. What is more problematic is how they are chosen. The pretrained LLM has likely \emph{never been trained to generate} tasks, from which stem the following observations:
\begin{itemize}
    \itemsep 0em
    \item Generated tasks are very \textbf{sensitive to the prompt}, leading to the need for prompt engineering. The agent does not decide to explore the environment by listing and reading files, unless this has been encouraged in the system prompt. If it is, it often gets stuck in a loop that reads the same files.
    \item They greatly \textbf{depend on long-term memory} (step 7. in Fig. \ref{fig: framework}). Sometimes the agent forgets to store information that a particular task has been done, which causes it to repeatedly generate the same task in the next run. It may also write down the task's result, but not the task itself, leading to the same outcome. Encouraging it to store a (task, action, outcome) tuple produces best results in terms of diverse tasks generated across multiple runs.
    \item Tasks follow the \textbf{statistical patterns} on which the LLM has been trained. When run for multiple runs without human supervision, the agent decided to implement a calculator, a password generator, a leap year checker, a prime number checker, a Celsius-Fahrenheit converter, a palindrome checker, multiple perfect and square number checkers. These are common tasks in the training data.
    \item Tasks are \textbf{steerable} through user feedback. We run the agent interactively and only occasionally provide user feedback, prompting it to choose certain tasks over others based on novelty. It responds by reasonably adjusting its generated tasks in the next run. However, since it does not decide to store the user's feedback, this adjustment is short-lived and the user's feedback is lost.
\end{itemize}

\begin{figure}[t]
    \centering
    \includegraphics[width=\columnwidth]{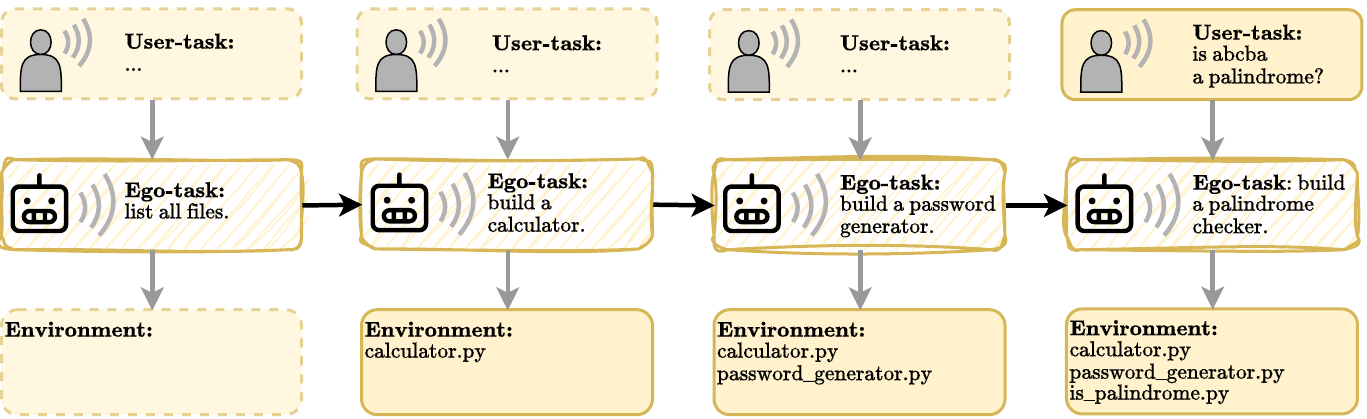}
    \caption{\textbf{A trajectory of sampled states.}
    The open-ended agent can generate tasks while incorporating user feedback. Its impact can be assessed through its artifacts in the environment.}
    \label{fig: chain_of_states}
\end{figure}
\section{Conclusion \& Future Work}
\label{sec: discussion}

Pretrained LLMs are optimized to be capable single-run problem solvers: they excel at well-defined, concrete natural language tasks, because that is what they were trained to do. Frameworks such as \texttt{ReAct} \cite{yao2023react} extend this ability through planning and tool use, broadening the scope of solvable problems, but fundamentally preserve the single-run, task-solving paradigm.

We tested such an agent in an open-ended environment where it must generate its own goals and sustain extended interactions. This setting raises new challenges: deciding which tasks to pursue, balancing novelty with continuity, building incrementally on prior goals, and selecting tasks of appropriate difficulty that are neither too hard, nor too easy. Current pretrained LLMs are not designed for this. The consequences are sensitivity to prompt design, repeated task generation, inadequate self-representation, and failure to identify what is worth storing long-term.

We conclude from these observations that LLMs could become strong open-ended agents if trained for such desired traits. For this reason, our future work will focus on directly training such agents to manage memory, explore productively, and select tasks that build toward abstract goal states. Similar to how techniques like GRPO \cite{guo2025deepseek} have been used to learn reasoning patterns for logical problem-solving \cite{shao2024deepseekmath}, one could just as well use them to learn action patterns for open-ended decision making. There is no reason why even these higher-order skills cannot be learned through experience.

\bibliographystyle{plain}
\bibliography{main.bib}

\end{document}